\documentclass[letterpaper]{article} 
\usepackage{aaai2026}  
\usepackage{times}  
\usepackage{helvet}  
\usepackage{courier}  
\usepackage[hyphens]{url}  
\usepackage{graphicx} 
\usepackage{booktabs}
\usepackage{graphicx}
\usepackage{amsmath}
\usepackage{amssymb}
\usepackage{multirow}
\usepackage{amsfonts}
\usepackage{booktabs}
\usepackage{graphicx}
\usepackage{natbib}  
\usepackage{caption} 
\usepackage{algorithm}
\usepackage{algorithmic}

\usepackage{newfloat}
\usepackage{listings}
\DeclareCaptionStyle{ruled}{labelfont=normalfont,labelsep=colon,strut=off} 
\floatstyle{ruled}
\newfloat{listing}{tb}{lst}{}
\floatname{listing}{Listing}
\title{DeViGrasp: Robust Visual Mobile Grasping for Quadruped Manipulators under Degraded Perception}
\author{
    Liang Zhou\textsuperscript{\rm 1},
    Jiaming Su\textsuperscript{\rm 2},
    Yancong Wei\textsuperscript{\rm 1},
    Kangkang Dong\textsuperscript{\rm 1},
    Houde Liu\textsuperscript{\rm 1}\thanks{Corresponding author.}
}

\affiliations{
    \textsuperscript{\rm 1}Tsinghua Shenzhen International Graduate School,
    Tsinghua University, Shenzhen, China\\
    \textsuperscript{\rm 2}Shanghai Jiao Tong University Global College,
    Shanghai Jiao Tong University, Shanghai, China\\
}

\usepackage{bibentry}
\nocopyright
\begin{document}

\maketitle

\begin{abstract}
Quadruped manipulators enable mobile grasping in complex environments, yet their whole-body control policies remain vulnerable to unreliable onboard visual perception. Existing methods are typically developed under relatively reliable observations and have not systematically examined how occlusion, segmentation-mask dropout, depth noise, and target-localization jitter affect grasp reasoning and target tracking. To address this gap, we introduce DeViGrasp-Bench, a benchmark for mobile grasping under degraded vision that incorporates controlled visual degradations, seen and unseen objects, multiple difficulty levels, and complex terrains, and evaluates task success, execution efficiency, and action smoothness. We further propose DeViGrasp-Net, a teacher--student framework that combines state-conditioned grasp reasoning with reliability-aware temporal target estimation. The privileged teacher attends to offline grasp candidates conditioned on object, robot, end-effector, and task states, while the deployable student fuses dual-view segmented-depth observations with current, memory, and recovery target hypotheses through Target Hold Memory and Temporal Memory Attention. DeViGrasp-Net outperforms VBC across degradation levels, unseen objects, and complex terrains, and surpasses an adapted DQ-Net across all evaluated degradation levels. Under the Difficult setting, it achieves a success rate of 62.3\%, improving upon VBC and DQ-Net by 16.1 and 4.3 percentage points, respectively; under the Hard setting, its margin over DQ-Net increases to 10.5 percentage points. Ablation studies confirm the complementary benefits of grasp-aware supervision and reliability-aware temporal memory.
\end{abstract}

\section{Introduction}

Quadruped manipulators combine legged mobility with arm dexterity, enabling mobile grasping in complex and unstructured environments~\cite{sleiman2021unified,liu2025visual,wang2025quadwbg,
zhang2025multistage,liang2026dqnet,fu2023deep,
ha2025umi,zhi2025unified}. Unlike tabletop manipulators, they must approach target objects while coordinating base motion, whole-body balance, arm control, and grasp execution. Moreover, their onboard cameras move continuously with the robot, making target observations vulnerable to viewpoint changes, environmental and self-occlusion, segmentation-mask dropout, depth noise, and localization jitter~\cite{miki2022perceptive,fang2023anygrasp,xia2026targo,
hu2023camera}. These perception errors can produce inaccurate target estimates, unstable end-effector motion, and inconsistent whole-body commands. Robust visual mobile grasping therefore requires both grasp reasoning compatible with the current whole-body state and stable target estimation under unreliable observations.

Existing legged mobile-manipulation methods mainly address whole-body coordination, task feasibility, or dynamic-object grasping under relatively reliable perception~\cite{liu2025visual,wang2025quadwbg,zhang2025multistage}. For example, DQ-Net employs privileged grasp memory and temporal dual-view encoding for dynamic-object grasping~\cite{liang2026dqnet}, but was not designed to explicitly estimate observation reliability or recover from intermittent mask and depth failures. Existing benchmarks also rarely characterize how controlled degradation types and severity levels affect grasping success, execution efficiency, and control stability~\cite{james2020rlbench,mu2021maniskill,
srivastava2022behavior,li2023behavior1k,
fang2020graspnet,xia2026targo,liu2023libero,
nasiriany2024robocasa,pumacay2024colosseum,
fei2026liberoplus}. Simple temporal smoothing may suppress high-frequency noise, but cannot selectively reject corrupted observations and may introduce tracking latency. A systematic evaluation framework and a reliability-aware policy are therefore both needed.

To address the evaluation gap, we introduce \textbf{DeViGrasp-Bench}, a degraded-vision mobile-grasping benchmark for quadruped manipulators. It applies controlled depth noise, segmentation-mask dropout, object occlusion, and target-localization jitter at four difficulty levels. The benchmark includes diverse seen and unseen objects~\cite{calli2015ycb}, randomized table heights, and flat, rough, sloped, and stair terrains. It evaluates task success, completion efficiency, and action smoothness, enabling systematic analysis of mobile-grasping robustness under combined visual and locomotion disturbances.

We further propose \textbf{DeViGrasp-Net}, a teacher--student framework combining state-conditioned grasp reasoning with reliability-aware temporal target estimation. The privileged teacher attends to offline grasp candidates~\cite{mahler2017dexnet,mousavian2019graspnet,sundermeyer2021contact} conditioned on object, robot, end-effector, and task states, producing a grasp representation compatible with the current whole-body configuration. The deployable student uses forward- and wrist-view segmented-depth observations together with proprioception. It constructs current, memory, and recovery target hypotheses through Target Hold Memory and recursive recovery, and fuses them according to their reliability using Temporal Memory Attention. The resulting stabilized representation predicts high-level base and arm commands executed through a hierarchical whole-body control interface.

Across four visual degradation levels, DeViGrasp-Net improves success rate and action smoothness over VBC, an EMA-based baseline, and an adapted DQ-Net~\cite{liu2025visual,liang2026dqnet}. Under the Difficult setting, it achieves a success rate of 62.3\%, exceeding VBC and DQ-Net by 16.1 and 4.3 percentage points, respectively. Under the Hard setting, its margin over DQ-Net further increases to 10.5 percentage points. DeViGrasp-Net also generalizes to unseen objects, complex terrains, and low-speed dynamic targets on DQ-Bench. Ablation studies confirm the complementary benefits of grasp-aware supervision and reliability-aware temporal memory.

Our main contributions are summarized as follows:
\begin{itemize}
    \item We introduce \textbf{DeViGrasp-Bench}, a benchmark for quadruped mobile grasping under controlled visual degradations, diverse objects, multiple difficulty levels, and complex terrains, with metrics covering task success, execution efficiency, and control stability.

    \item We propose \textbf{DeViGrasp-Net}, which combines state-conditioned grasp attention in a privileged teacher with reliability-aware temporal target estimation in a deployable student.

    \item We conduct systematic experiments across degradation
severity, unseen objects, complex terrains, and low-speed dynamic
targets, with ablations validating the proposed components.
\end{itemize}

\begin{figure*}[t]
    \centering
    \includegraphics[width=\textwidth]{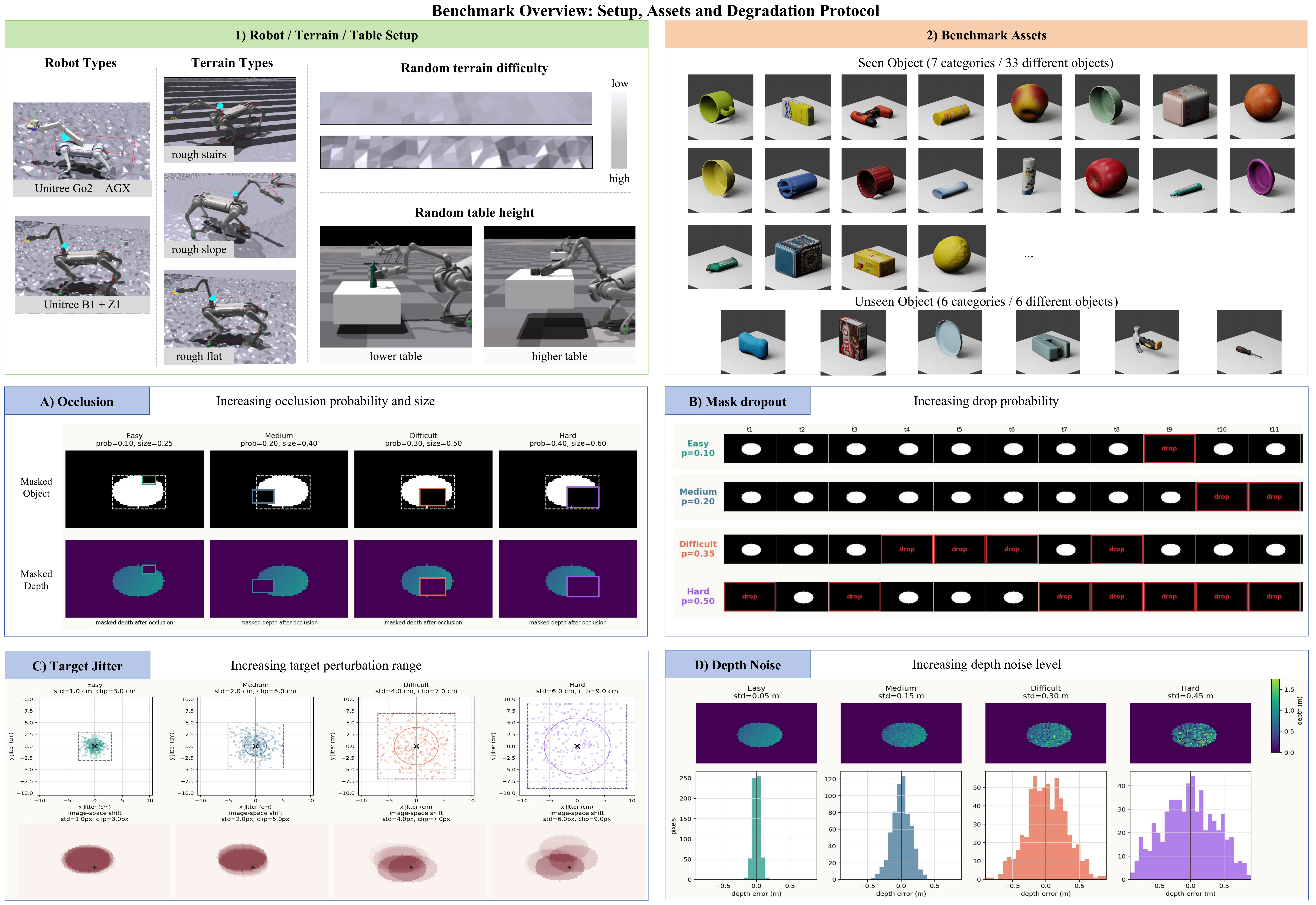}
\caption{
Overview of \textbf{DeViGrasp-Bench}. 
It evaluates quadruped mobile grasping across diverse objects, terrain conditions, and visual degradation levels, including depth noise, mask dropout, occlusion, and target localization jitter.
}
    \label{fig:benchmark}
\end{figure*}

\section{Related Work}

\subsection{Legged Mobile Manipulation}

Legged mobile manipulators combine terrain-adaptive locomotion with arm-based interaction~\cite{sleiman2021unified,liu2025visual,wang2025quadwbg,
fu2023deep,ha2025umi,zhi2025unified}. Optimization-based methods coordinate locomotion and manipulation through whole-body control or model predictive control~\cite{sleiman2021unified,yao2022transferable,
risiglione2022wholebody}, while learning-based approaches train policies for leg--arm coordination~\cite{rudin2022walk,miki2022perceptive,
fu2023deep,ha2025umi,zhi2025unified}. VBC~\cite{liu2025visual} adopts a hierarchical visual whole-body controller for quadruped mobile grasping, and later methods further improve grasp reasoning and task generalization~\cite{zhang2024gamma,wang2025quadwbg,
zhang2025multistage,ha2025umi}. For dynamic-object grasping, DQ-Net~\cite{liang2026dqnet} uses a
privileged grasp memory and temporal dual-view encoding. However, these methods mainly study control coordination or target motion under relatively reliable observations. In contrast, DeViGrasp focuses on intermittent perception failures and explicitly models grasp compatibility and visual-target reliability through state-conditioned grasp attention and current, memory, and recovery target hypotheses.

\subsection{Robust Manipulation Benchmarks}

Existing benchmarks evaluate visual manipulation from complementary perspectives~\cite{yu2020metaworld,xiang2020sapien,
ehsani2021manipulathor,mees2022calvin,
dasari2020robonet,liu2023libero,
nasiriany2024robocasa,pumacay2024colosseum,
fei2026liberoplus}. RLBench~\cite{james2020rlbench} and ManiSkill~\cite{mu2021maniskill} emphasize task diversity and object generalization, while GraspNet~\cite{fang2020graspnet} focuses on 6-DoF grasp detection and TARGO~\cite{xia2026targo} studies target grasping under occlusion. Locomotion benchmarks instead evaluate terrain traversal and control robustness~\cite{rudin2022walk,miki2022perceptive}. However, these benchmarks do not jointly examine quadruped whole-body grasping, complex terrain, and multiple controlled visual degradations. DeViGrasp-Bench addresses this gap by evaluating mask dropout, object occlusion, depth noise, and target-localization jitter at multiple severity levels, together with seen/unseen objects, terrain variation, task efficiency, and action smoothness.

\begin{table}[t]
\centering
\footnotesize
\setlength{\tabcolsep}{2pt}
\begin{tabular}{@{}lcccccc@{}}
\toprule
\textbf{Level} &
\textbf{Depth} &
\textbf{Dropout} &
\textbf{Occ. P.} &
\textbf{Occ. S.} &
\textbf{Jitter Std} &
\textbf{Jitter Clip} \\
\midrule
Easy      & 0.05 & 0.10 & 0.10 & 0.25 & 0.01 & 0.03 \\
Medium    & 0.15 & 0.20 & 0.20 & 0.40 & 0.02 & 0.05 \\
Difficult & 0.30 & 0.35 & 0.30 & 0.50 & 0.04 & 0.07 \\
Hard      & 0.45 & 0.50 & 0.40 & 0.60 & 0.06 & 0.09 \\
\bottomrule
\end{tabular}
\caption{Visual degradation levels in DeViGrasp-Bench.}
\label{tab:visual_degradation_levels}
\end{table}

\section{DeViGrasp-Bench: Degraded Vision Grasping Benchmark}

Quadruped manipulators rely on moving onboard cameras to observe target objects, making visual perception vulnerable to environmental occlusion, arm self-occlusion, segmentation-mask dropout, depth noise, and target-localization errors. To systematically study how these failures affect closed-loop mobile grasping, we introduce \textbf{DeViGrasp-Bench}, a degraded-vision grasping benchmark for quadruped manipulators. An overview of the benchmark is shown in Fig.~\ref{fig:benchmark}.

\paragraph{Task and Assets.}
DeViGrasp-Bench is built in Isaac Gym. At the beginning of each episode, a target object is placed on a table with randomized height, and the robot must approach the workspace, align its base and end-effector, grasp the target, and maintain whole-body stability. A trial is considered successful when the object is lifted by more than 0.35\,m above its initial height and remains lifted for 25 high-level control steps. The benchmark contains 33 seen object instances for training and standard evaluation and 6 unseen object instances used only for generalization testing. These objects cover diverse geometries, sizes, and grasp affordances, including bowls, balls, boxes, bottles, cups, and elongated tool-like objects. The benchmark supports flat ground and complex terrains, including rough flat terrain, slopes, and stairs. All quantitative experiments use a Unitree B1 quadruped equipped with a Unitree Z1 arm. We additionally provide qualitative demonstrations on Unitree Go2 in the supplementary material.

\paragraph{Visual Degradation.}
At each time step, controlled perturbations are independently applied to the forward- and wrist-view perception streams. Depth noise adds Gaussian perturbations to valid depth pixels. Mask dropout removes the complete target mask and its corresponding segmented-depth observation. Object occlusion removes a local region from both the target mask and segmented depth. Target-localization jitter adds zero-mean Gaussian perturbations to the estimated 3D target position, where the jitter standard deviation is specified in meters for each coordinate axis and the perturbation is bounded by a clipping range. We define four composite difficulty levels---Easy, Medium, Difficult, and Hard---with progressively stronger degradation parameters, as summarized in Table~\ref{tab:visual_degradation_levels}. These controlled operators are designed to isolate the robustness of the downstream target-estimation and control pipeline from the accuracy of any particular segmentation or detection model.

\paragraph{Metrics.}
DeViGrasp-Bench evaluates policies in terms of task success, execution efficiency, and control stability. We report Success Rate (SR), defined as the percentage of successful episodes; Average Episode Steps (AES), the mean episode length over all trials; Average Steps to Success (ASTS), the mean number of steps among successful trials; and Action Smoothness (AS-L1), the average L1 difference between consecutive high-level actions. Together, these metrics characterize grasping reliability, completion efficiency, and whole-body command variation under degraded visual perception. Additional details on object splits, terrain generation and evaluation protocols are provided in the supplementary material.

\begin{figure*}[t]
    \centering
    \includegraphics[width=\textwidth]{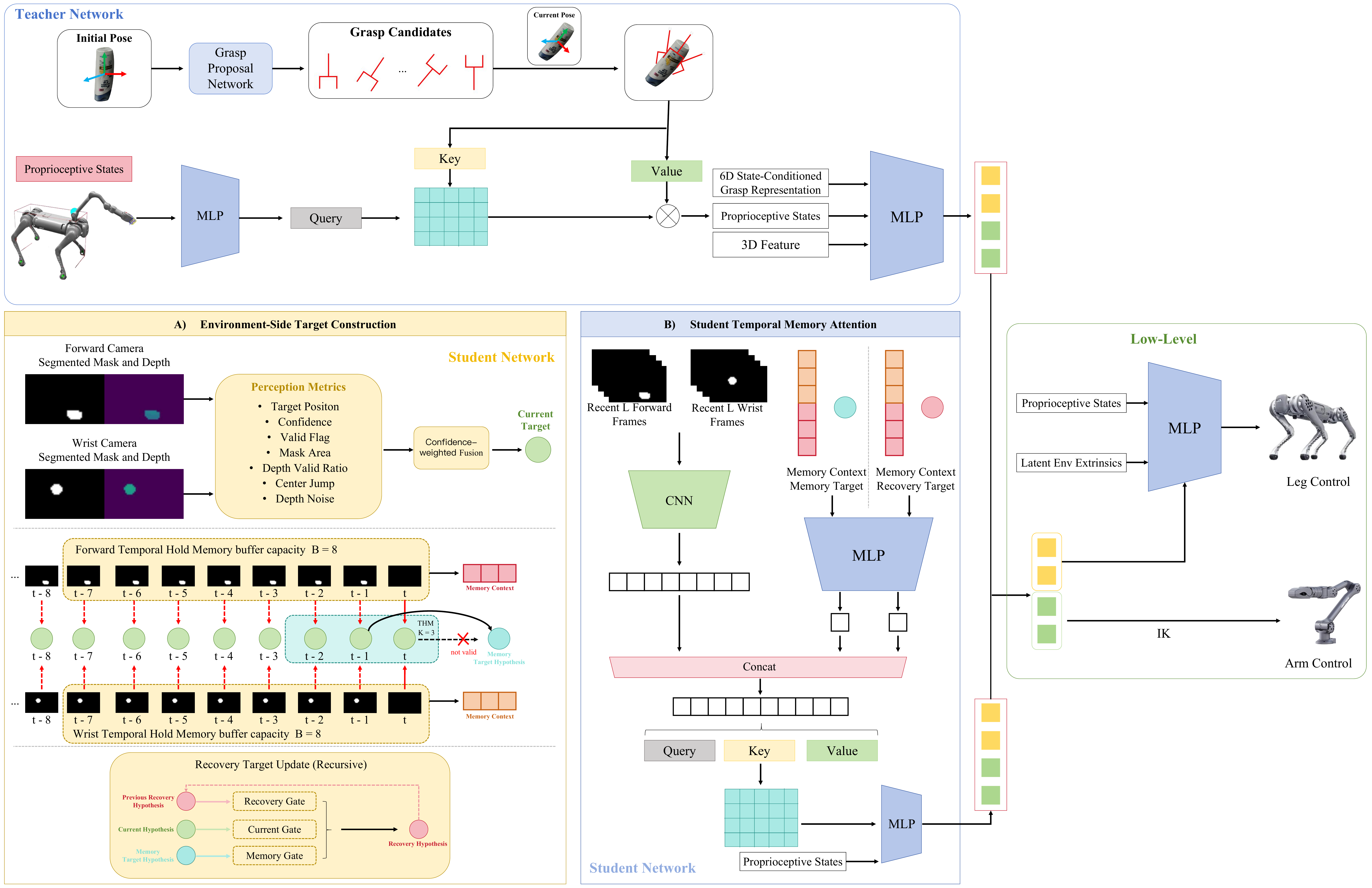}
\caption{
Overview of \textbf{DeViGrasp-Net}. 
A privileged teacher learns grasp-aware whole-body actions from object, robot, task, and grasp-candidate information, while a deployable student uses forward/wrist segmented depth observations and temporal memory attention for robust target estimation under degraded perception.
}
    \label{fig:method_overview}
\end{figure*}

\section{Method}

\subsection{Overview}

We propose \textbf{DeViGrasp-Net}, a teacher--student framework for
robust quadruped mobile grasping under degraded visual perception.
As illustrated in Fig.~\ref{fig:method_overview}, the framework
addresses two complementary challenges: selecting grasp candidates
that match the current whole-body state and maintaining a reliable
target representation under intermittent visual failures.

At each high-level control step, both policies predict
\begin{equation}
\mathbf{u}_t=
[\Delta\mathbf{p}_t,\Delta\mathbf{r}_t,g_t,v_t,\omega_t]
\in\mathbb{R}^{9},
\end{equation}
where $\Delta\mathbf{p}_t$ and $\Delta\mathbf{r}_t$ are incremental
end-effector position and RPY-orientation commands, $g_t$ is the
gripper command, and $v_t$ and $\omega_t$ are the base forward and
yaw velocity commands. The privileged teacher and deployable student
operate on privileged and visual observations, respectively:
\begin{equation}
\mathbf{u}^{T}_t=\pi_T(\mathbf{o}^{T}_t), \qquad
\mathbf{u}^{S}_t=\pi_S(\mathbf{o}^{S}_t).
\end{equation}

\subsection{Grasp-Aware Privileged Teacher}

The teacher receives an object feature $\mathbf{f}_o$, robot state
$\mathbf{s}_t$, end-effector state $\mathbf{e}_t$, task context
$\boldsymbol{\xi}_t$, and $N$ offline grasp candidates
$\mathcal{G}_t=\{\mathbf{g}_t^i\}_{i=1}^{N}$, where
$\mathbf{g}_t^i\in\mathbb{R}^{6}$ contains position and RPY
orientation in an arm-base frame aligned with the robot body. The
candidates are generated offline for each object and transformed online
into the current arm-base frame at every high-level step. The attention
module and teacher policy are optimized jointly through PPO. The attention
variables are constructed as
\begin{equation}
\begin{aligned}
\mathbf{q}_t &=W_q[\phi_o(\mathbf{f}_o),\mathbf{s}_t,
\mathbf{e}_t,\boldsymbol{\xi}_t],\\
\mathbf{k}_t^i&=W_k\mathbf{g}_t^i, \qquad
\boldsymbol{\nu}_t^i=W_v\mathbf{g}_t^i,\\
\alpha_t^i&=
\frac{\exp(\mathbf{q}_t^\top\mathbf{k}_t^i/\sqrt d)}
{\sum_j\exp(\mathbf{q}_t^\top\mathbf{k}_t^j/\sqrt d)}.
\end{aligned}
\end{equation}
The resulting state-conditioned grasp context is
\begin{equation}
\mathbf{z}^{g}_t =
W_o \sum_i \alpha_t^i \boldsymbol{\nu}_t^i,
\qquad
\mathbf{u}^{T}_t =
\pi_T([\phi_o(\mathbf{f}_o), \mathbf{s}_t,
\mathbf{e}_t, \boldsymbol{\xi}_t, \mathbf{z}^{g}_t]).
\end{equation}
Here, $\mathbf{z}^{g}_t \in \mathbb{R}^{d_g}$ is a latent grasp-context
embedding used to condition the teacher policy. The attention module
aggregates projected grasp features rather than directly averaging
candidate poses in the geometric pose space.

\subsection{Visual Student with Target Hold Memory and Temporal Memory Attention}

For each view $v\in\{f,w\}$, denoting the forward and wrist cameras,
the student receives a segmented mask--depth observation
$\mathbf{I}_t^v$. We extract a normalized image--depth descriptor and
a deterministic reliability score:
\begin{equation}
\boldsymbol{\tau}_t^v=
\left[u_t^v/W,\;v_t^v/H_I,\;\bar d_t^v\right],
\qquad
\rho_t^v=\mathcal{R}(\mathbf{I}_t^v),
\end{equation}
where $(u_t^v,v_t^v)$ is the mask center, $\bar d_t^v$ is the
normalized mean target depth, and $\mathcal{R}$ evaluates mask area,
valid-depth ratio, center displacement, and depth variation. These
camera-specific descriptors are policy features rather than coordinates
in a shared geometric frame. Their reliability-weighted summary is
\begin{equation}
\boldsymbol{\tau}^{\mathrm{cur}}_t=
\frac{\rho_t^f\boldsymbol{\tau}_t^f+
\rho_t^w\boldsymbol{\tau}_t^w}
{\rho_t^f+\rho_t^w+\epsilon},
\end{equation}
where $\epsilon>0$ ensures numerical stability. If both views are
invalid, the previous recovery descriptor is used and the current gate
is set to zero. Both this summary and the TMA-pooled descriptor in Eq.~(9) are policy
features, not metric 3D estimates. TMA retains separate per-view
tokens with view-specific projections and source embeddings. Thus, the
fusion assumes no geometric or pixel-wise correspondence between
cameras.

\paragraph{Target Hold Memory.}
Target Hold Memory (THM) maintains a per-view buffer
$\mathcal{B}_t^v$ with capacity $B$ for recent valid masks and
segmented-depth observations. When the current target is missing, THM
retrieves the most recent valid observation whose age does not exceed
the hold horizon $K$, producing $\widetilde{\mathbf{I}}_t^v$ and the
visual feature $\mathbf{z}_t^v=\phi_v(\widetilde{\mathbf{I}}_t^v)$.
It also maintains a reliable descriptor
$\boldsymbol{\tau}_t^{\mathrm{mem}}$. This descriptor provides a
short-term fallback bounded by $K$. In contrast, a persistent recovery
descriptor remains available after the hold horizon expires and
recursively aggregates the current, memory, and previous recovery
hypotheses:
\begin{equation}
\boldsymbol{\tau}^{\mathrm{rec}}_t=
\lambda_t^{\mathrm{cur}}\boldsymbol{\tau}^{\mathrm{cur}}_t+
\lambda_t^{\mathrm{mem}}\boldsymbol{\tau}^{\mathrm{mem}}_t+
\lambda_t^{\mathrm{rec}}\boldsymbol{\tau}^{\mathrm{rec}}_{t-1},
\end{equation}
where the normalized gates depend on current reliability, memory
validity, and freshness.

\paragraph{Temporal Memory Attention.}
Temporal Memory Attention (TMA) fuses target tokens from the most
recent $L$ high-level steps of each camera with the memory and recovery
tokens:
\begin{equation}
\mathcal{T}_t=
\{\mathbf{T}_{t-L+1:t}^{f},\mathbf{T}_{t-L+1:t}^{w},
\mathbf{T}_t^{\mathrm{mem}},\mathbf{T}_t^{\mathrm{rec}}\}.
\end{equation}
Each camera token contains its target descriptor, reliability, validity,
and perception statistics. The memory and recovery tokens combine their
descriptors with confidence, validity, age, and freshness context.
Source-specific projections and temporal embeddings preserve camera
identity and temporal order, while invalid tokens are masked. A
Transformer encoder produces the stabilized descriptor and temporal
representation:
\begin{equation}
\begin{aligned}
\mathbf{Z}_t&=\operatorname{Transformer}(\mathcal{T}_t), \qquad
\beta_t^i=\operatorname{softmax}_i
(\mathbf{w}_\beta^\top\mathbf{Z}_t^i),\\
\hat{\boldsymbol{\tau}}_t&=\sum_i\beta_t^i\boldsymbol{\tau}_t^i,
\qquad
\mathbf{m}_t=\sum_i\beta_t^i\mathbf{Z}_t^i.
\end{aligned}
\end{equation}
The student combines these outputs with visual features,
proprioception, task context, and source-level attention weights
$\boldsymbol{\gamma}_t$, which summarize the contributions of the
forward view, wrist view, memory target, and recovery target. A GRU-based
recurrent policy predicts
\begin{equation}
\begin{aligned}
\mathbf{x}_t^S&=[\mathbf{z}_t^f,\mathbf{z}_t^w,
\hat{\boldsymbol{\tau}}_t,\mathbf{s}_t,
\boldsymbol{\xi}_t,\mathbf{m}_t,\boldsymbol{\gamma}_t],\\
\mathbf{h}_t&=\operatorname{GRU}(\mathbf{x}_t^S,\mathbf{h}_{t-1}),
\qquad \mathbf{u}^{S}_t=\psi(\mathbf{h}_t).
\end{aligned}
\end{equation}
We use $B=8$ observations per view, a hold horizon of $K=3$ high-level
steps, and an attention window of $L=4$ steps per camera.

\subsection{Hybrid Low-Level Control}

The high-level command is executed through locomotion, arm, and
gripper branches. A pretrained locomotion policy uses proprioception,
base commands, and the current end-effector target to produce the
12-dimensional leg command
\begin{equation}
\mathbf{a}^{\mathrm{leg}}_t=
[\pi_L(\mathbf{o}_t^L,v_t,\omega_t,
\mathbf{y}_t^{ee})]_{1:12}.
\end{equation}
The end-effector command is tracked by a damped least-squares
inverse-kinematics controller, while $g_t$ is passed to the gripper
controller. Thus, locomotion and arm control remain separate while
sharing the high-level task command.

\subsection{Training}

The privileged teacher is first trained using PPO. The visual student is
then trained under degraded observations using a DAgger-style on-policy
imitation procedure, with the teacher providing labels on states visited
by the student. The student minimizes
\begin{equation}
\mathcal{L}_{\mathrm{BC}}=
\mathbb{E}_t\left[
\|\mathbf{u}^{S}_t-\mathbf{u}^{T}_t\|_2^2
\right].
\end{equation}
Depth noise, mask dropout, object occlusion, and target-localization
jitter are injected during student training. Additional architecture,
update-rule, control, and training details are provided in
supplementary material.

\begin{figure}[t]
    \centering
    \includegraphics[width=\columnwidth]{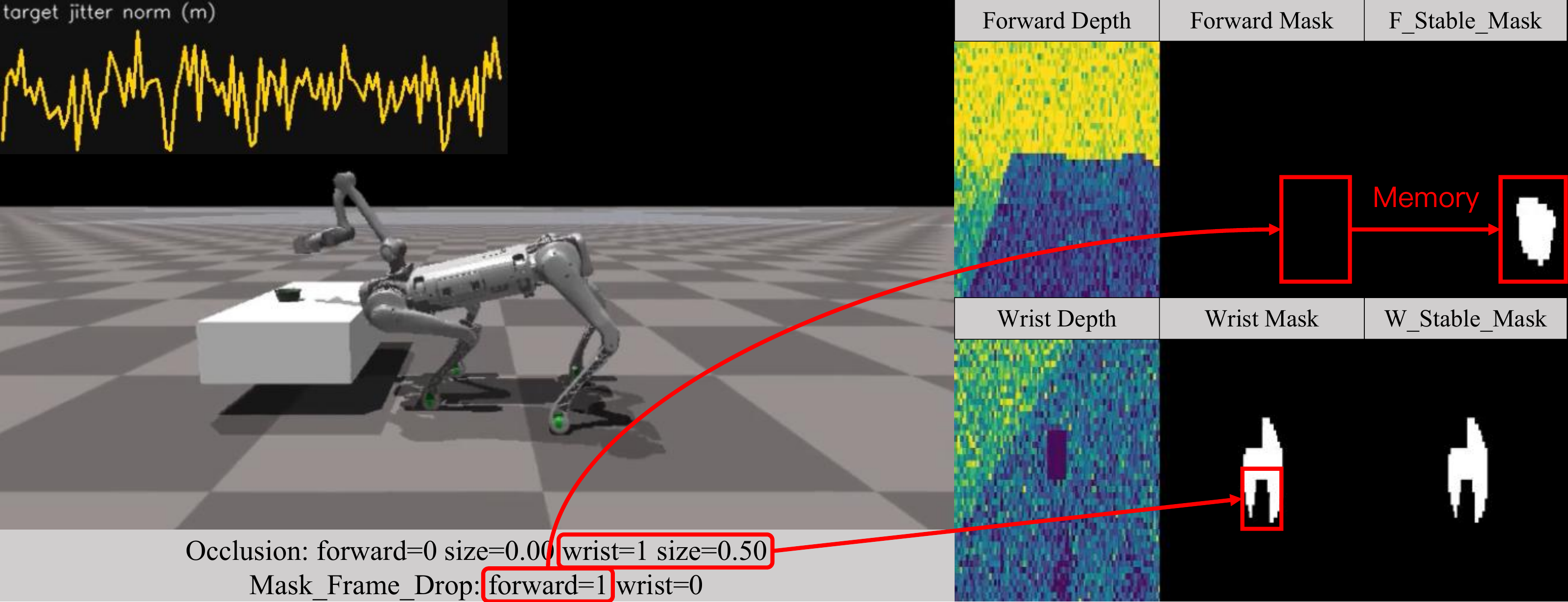}
\caption{
Visualization of Target Hold Memory under mask
dropout and partial occlusion. Memory-retained observations
preserve usable target evidence when current observations are
corrupted.
}
    \label{fig:memory_vis}
\end{figure}

\begin{figure*}[t]
    \centering
    \includegraphics[width=\textwidth]{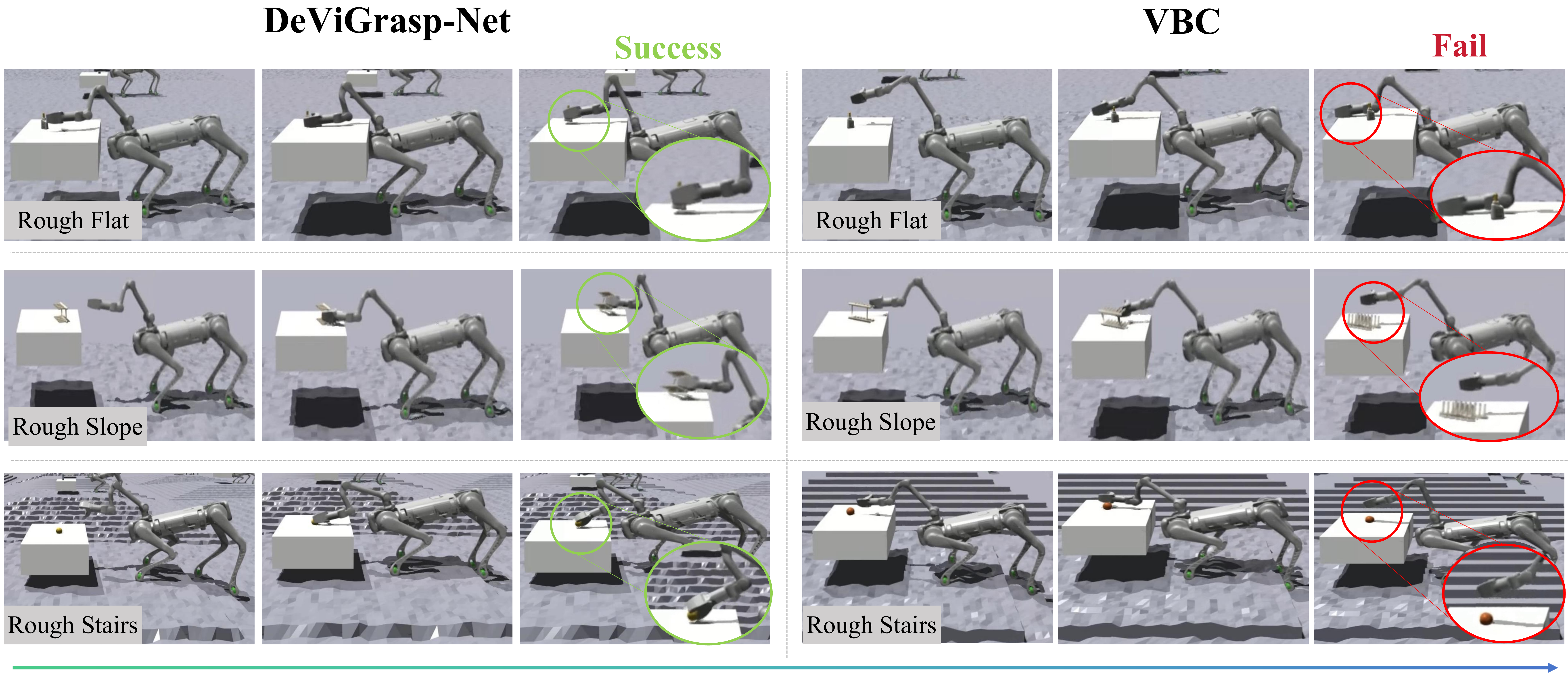}
\caption{
Qualitative comparison on complex terrains. 
DeViGrasp-Net achieves stable grasps across rough flat, slope, and stairs, while VBC often fails from target-tracking and end-effector misalignment.
}
    \label{fig:overview3}
\end{figure*}

\begin{table*}[t]
\centering
\footnotesize
\setlength{\tabcolsep}{0.8pt}

\begin{tabular}{@{}l*{16}{c}@{}}
\toprule
\multirow{2}{*}{\textbf{Model}}
& \multicolumn{4}{c}{\textbf{Easy}}
& \multicolumn{4}{c}{\textbf{Medium}}
& \multicolumn{4}{c}{\textbf{Difficult}}
& \multicolumn{4}{c}{\textbf{Hard}} \\
\cmidrule(lr){2-5}
\cmidrule(lr){6-9}
\cmidrule(lr){10-13}
\cmidrule(lr){14-17}

& \textbf{AES} $\downarrow$
& \textbf{ASTS} $\downarrow$
& \textbf{AS-L1} $\downarrow$
& \textbf{SR} $\uparrow$

& \textbf{AES} $\downarrow$
& \textbf{ASTS} $\downarrow$
& \textbf{AS-L1} $\downarrow$
& \textbf{SR} $\uparrow$

& \textbf{AES} $\downarrow$
& \textbf{ASTS} $\downarrow$
& \textbf{AS-L1} $\downarrow$
& \textbf{SR} $\uparrow$

& \textbf{AES} $\downarrow$
& \textbf{ASTS} $\downarrow$
& \textbf{AS-L1} $\downarrow$
& \textbf{SR} $\uparrow$ \\
\midrule

VBC
& 100.9 & 61.2 & 0.372 & 43.0
& 93.6  & 67.3 & 0.370 & 47.2
& 94.9  & 66.2 & 0.370 & 46.2
& 108.3 & 79.7 & 0.372 & 29.3 \\

VBC+EMA
& 91.4  & 62.8 & 0.366 & 47.8
& 98.5  & 64.8 & 0.365 & 38.5
& 104.2 & 70.3 & 0.361 & 37.9
& 107.6 & 74.7 & 0.362 & 28.6 \\

DQ-Net
& \textbf{74.4} & \textbf{51.8} & 0.339 & 64.5
& \textbf{74.8} & \textbf{52.7} & 0.339 & 66.2
& \textbf{79.5} & \textbf{55.1} & 0.336 & 58.0
& 93.8 & \textbf{61.1} & 0.337 & 44.3 \\

\textbf{DeViGrasp-Net}
& 75.3 & 54.8 & \textbf{0.291} & \textbf{70.2}
& 76.4 & 55.0 & \textbf{0.290} & \textbf{68.3}
& 82.6 & 58.3 & \textbf{0.289} & \textbf{62.3}
& \textbf{88.1} & 61.6 & \textbf{0.287} & \textbf{54.8} \\

\bottomrule
\end{tabular}

\caption{
Main results under different visual degradation levels on flat terrain.
Results are averaged over three random seeds; full mean$\pm$std
statistics are provided in the supplementary material.
}
\label{tab:main_visual_degradation}
\end{table*}

\section{Experiments}

We evaluate \textbf{DeViGrasp-Net} on \textbf{DeViGrasp-Bench}. 
The experiments are designed to answer four questions: 
(1) whether DeViGrasp-Net improves robustness under different levels of visual degradation; 
(2) how each proposed component contributes to performance; 
(3) whether the learned policy generalizes to unseen objects; and 
(4) whether the method remains effective on complex terrain.

\subsection{Experimental Setup}

All experiments use Isaac Gym on a single NVIDIA RTX 3090 Ti
with a Unitree B1--Z1 platform. Methods share the high-level
command space, hybrid controller, and terrain-specific low-level
checkpoints. The teacher is trained for 60,000 high-level steps
using 10,240 environments and 24-step rollouts. VBC, VBC+EMA,
and DeViGrasp-Net use 240 camera environments for 60,000 steps
(14.4M transitions), whereas DQ-Net uses 200 environments for
80,000 steps (16M transitions). For DeViGrasp-Bench, all students
are trained under Difficult degradation and evaluated at all four
degradation levels. For the separate DQ-Bench study, DeViGrasp-Net
and DQ-Net are trained on Level~4 and evaluated on Level~1 without
level-specific fine-tuning, following the released task and control
settings.

For the main visual-degradation, ablation, and complex-terrain
experiments, results are averaged over 500 episodes per seed and
three seeds. The DQ-Bench results are also averaged over three
seeds. VBC excludes grasp attention and temporal memory; VBC+EMA
adds finite-horizon EMA smoothing; adapted DQ-Net retains its
grasp-memory teacher and temporal dual-view student under the same
degradation and evaluation protocol. Ablations isolate grasp
attention, THM, and TMA. Further details and seed-level statistics
are provided in the supplementary material.

\begin{table}[t]
\centering
\footnotesize
\setlength{\tabcolsep}{3pt}

\begin{tabular}{@{}lcccc@{}}
\toprule
\textbf{Setting}
& \textbf{AES} $\downarrow$
& \textbf{ASTS} $\downarrow$
& \textbf{AS-L1} $\downarrow$
& \textbf{SR (\%)} $\uparrow$ \\
\midrule
VBC
& 94.9 & 66.2 & 0.370 & 46.2 \\
w/ GA
& 89.1 & 64.8 & \textbf{0.278} & 53.5 \\
w/ GA + THM
& 85.5 & 66.2 & 0.282 & 57.1 \\
\textbf{w/ GA + THM + TMA}
& \textbf{82.6} & \textbf{58.3} & 0.289 & \textbf{62.3} \\
\bottomrule
\end{tabular}

\caption{
Ablation study under the Difficult visual degradation level.
Results are averaged over three random seeds.
}
\label{tab:ablation}
\end{table}

\begin{table}[t]
\centering
\footnotesize
\setlength{\tabcolsep}{2.3pt}

\begin{tabular}{@{}llcccc@{}}
\toprule
\textbf{Model}
& \textbf{Terrain}
& \textbf{AES} $\downarrow$
& \textbf{ASTS} $\downarrow$
& \textbf{AS-L1} $\downarrow$
& \textbf{SR (\%)} $\uparrow$ \\
\midrule
\multirow{3}{*}{VBC}
& R. Flat   & 130.6 & 81.6 & 0.311 & 7.9 \\
& R. Slope  & 131.5 & 82.9 & 0.311 & 6.4 \\
& R. Stairs & 130.3 & 66.5 & 0.328 & 10.1 \\
\midrule
\multirow{3}{*}{\textbf{DeViGrasp-Net}}
& R. Flat   & 92.1  & 64.1 & 0.290 & 58.4 \\
& R. Slope  & 89.1  & 65.7 & 0.291 & 61.5 \\
& R. Stairs & 111.8 & 65.7 & 0.293 & 34.2 \\
\bottomrule
\end{tabular}

\caption{
Complex-terrain evaluation under Difficult visual degradation.
Results are averaged over three random seeds.
}
\label{tab:complex_terrain}
\end{table}

\begin{table}[t]
\centering
\footnotesize
\setlength{\tabcolsep}{13pt}
\begin{tabular}{lccc}
\toprule
\textbf{Method}
& \textbf{GSR-T}
& \textbf{GSR-S}
& \textbf{OSSR} \\
\midrule
DQ-Net
& 80.8 & 55.8 & 53.2 \\
\textbf{DeViGrasp-Net}
&\textbf{85.2} & \textbf{60.7} & \textbf{57.6} \\
\bottomrule
\end{tabular}
\caption{Generalization to low-speed dynamic targets on DQ-Bench
Level~1. GSR-T and GSR-S denote teacher and student grasp success
rates, respectively, while OSSR denotes the one-shot success rate.
Results are averaged over three random seeds.}
\label{tab:dqbench_level1}
\end{table}

\subsection{Experiment Results}

We evaluate DeViGrasp-Net from four complementary aspects:
robustness across visual degradation levels, component
effectiveness, generalization to unseen object instances, and
robustness on complex terrains.

\paragraph{Robustness under visual degradation.}
Performance under progressively severe visual degradation demonstrates
consistent robustness and control-stability benefits for DeViGrasp-Net
(Table~\ref{tab:main_visual_degradation}). It achieves the highest SR
and the lowest AS-L1 across all difficulty levels. Compared with VBC,
DeViGrasp-Net improves SR by 16.1 and 25.5 percentage points under
Difficult and Hard, respectively. Relative to DQ-Net, it improves SR
by 4.3 and 10.5 percentage points in the same settings while reducing
AS-L1 by approximately 14\%. Although DQ-Net completes successful
episodes in fewer steps in several settings, DeViGrasp-Net consistently
achieves higher SR and lower AS-L1. VBC+EMA does not consistently
improve over VBC, suggesting that fixed temporal smoothing is
insufficient for intermittent and heterogeneous perception failures.
The largest SR margin under Hard further indicates that
reliability-aware target retention and temporal fusion become
particularly beneficial when visual observations are severely
corrupted.

\paragraph{Ablation study.}
Table~\ref{tab:ablation} evaluates the proposed components under the Difficult setting, with results averaged over three random seeds. Adding grasp-aware teacher supervision improves SR from 46.2\% to 53.5\% and reduces AS-L1 from 0.370 to 0.278, indicating more reliable and smoother whole-body control. Incorporating Target Hold Memory further increases SR to 57.1\%, demonstrating the benefit of retaining recent reliable observations during temporary perception failures. The full DeViGrasp-Net achieves 62.3\% SR and the lowest ASTS of 58.3 steps. These results demonstrate the complementary contributions of grasp-aware supervision, short-term target retention, and reliability-aware temporal fusion.


\begin{figure}[t]
    \centering
    \includegraphics[width=\columnwidth]{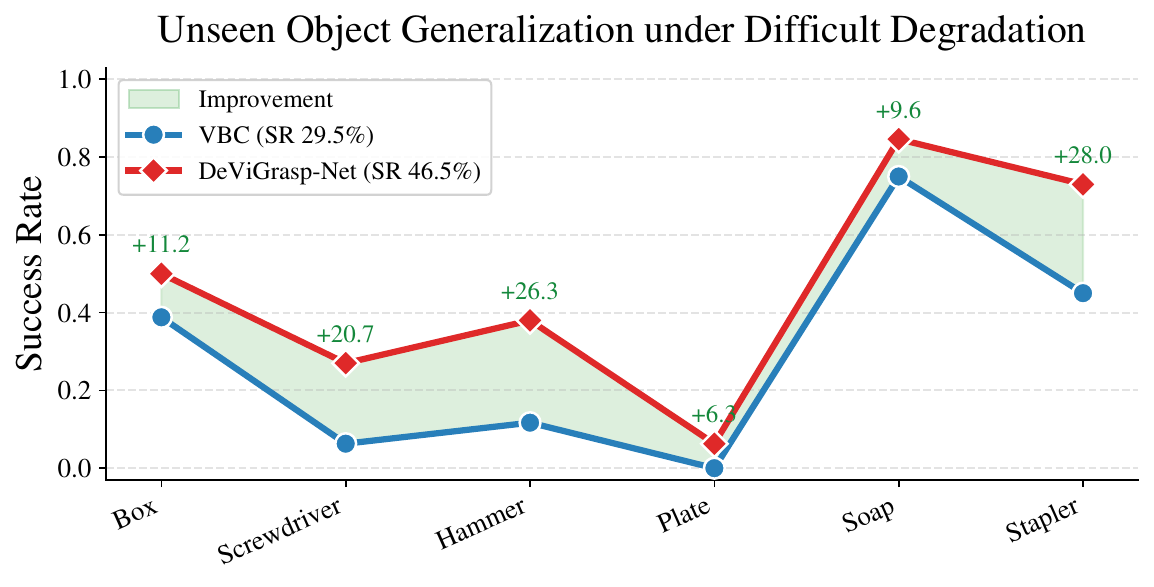}
\caption{Generalization to unseen object instances under the
Difficult degradation setting.}
    \label{fig:unseen_generalization}
\end{figure}

\paragraph{Generalization to unseen objects.}
We directly evaluate the trained policies on six unseen
object instances under the Difficult setting without
fine-tuning. As shown in Fig.~\ref{fig:unseen_generalization}, VBC
achieves 29.5\% aggregate SR, whereas DeViGrasp-Net
achieves 46.5\%. Improvements are observed across all
evaluated objects, suggesting that state-conditioned grasp
supervision and stable target representations generalize
beyond the training instances.

\paragraph{Complex-terrain evaluation.}
We further evaluate the policies on rough flat, rough slope,
and rough stair terrains under the Difficult visual degradation
setting. This evaluation requires simultaneous target tracking,
base locomotion, end-effector control, and whole-body
stability. As shown in Table~\ref{tab:complex_terrain},
DeViGrasp-Net achieves 58.4\%, 61.5\%, and 34.2\% SR on
rough flat, rough slope, and rough stair terrains, respectively,
while VBC achieves only 6.4--10.1\% SR. The lower
performance on stairs shows that combined locomotion and
visual disturbances remain challenging, but DeViGrasp-Net
consistently improves robustness across all terrain types.

\paragraph{Generalization to low-speed dynamic targets.} We evaluate 
DeViGrasp-Net on DQ-Bench Level~1. As shown in
Table~\ref{tab:dqbench_level1}, it achieves teacher and student grasp
success rates (GSR-T/GSR-S) of 85.2\%/60.7\% and a one-shot success
rate (OSSR) of 57.6\%, exceeding DQ-Net by 4.4, 4.9, and 4.4
percentage points, respectively. This indicates that reliability-aware
memory preserves responsiveness and policy transfer under low-speed
target motion.

\paragraph{Qualitative analysis.}
Fig.~\ref{fig:memory_vis} shows that Target Hold Memory retains usable
target evidence during mask dropout and partial occlusion. Under
combined terrain and visual disturbances, DeViGrasp-Net also maintains
steadier target tracking and end-effector alignment than VBC, enabling
more stable grasp attempts (Fig.~\ref{fig:overview3}).

\section{Conclusion}

We introduced \textbf{DeViGrasp-Bench}, a benchmark for
quadruped mobile grasping under controlled visual degradations,
diverse objects, and complex terrains. We further proposed
\textbf{DeViGrasp-Net}, a teacher--student framework combining
state-conditioned grasp attention in the privileged teacher with
reliability-aware temporal target estimation in the deployable
student. Across four degradation levels, DeViGrasp-Net achieves
higher success rates and smoother actions than VBC, an EMA-based
baseline, and an adapted DQ-Net. It also generalizes better than VBC to unseen objects and complex terrains, while outperforming DQ-Net on low-speed dynamic targets
in DQ-Bench. Ablations verify the
complementary contributions of grasp-aware supervision, short-term
target retention, and reliability-aware temporal fusion. These
results highlight the importance of jointly reasoning about grasp
compatibility and visual reliability for robust legged mobile
manipulation. Future work will investigate real-world deployment
and sim-to-real transfer under naturally occurring perception
failures.

\bibliography{main}

\end{document}